\documentclass{article} 
\usepackage{iclr2027_conference,times}

\usepackage{amsmath,amsfonts,bm}

\def\eqref#1{equation~\ref{#1}}

\def\1{\bm{1}}

\def\vg{{\bm{g}}}

\def\vk{{\bm{k}}}

\def\vo{{\bm{o}}}

\def\vq{{\bm{q}}}

\def\vv{{\bm{v}}}

\def\mA{{\bm{A}}}

\def\mI{{\bm{I}}}

\def\mR{{\bm{R}}}
\def\mS{{\bm{S}}}
\def\mT{{\bm{T}}}

\def\mLambda{{\bm{\Lambda}}}

\DeclareMathAlphabet{\mathsfit}{\encodingdefault}{\sfdefault}{m}{sl}
\SetMathAlphabet{\mathsfit}{bold}{\encodingdefault}{\sfdefault}{bx}{n}

\usepackage{hyperref}
\usepackage{url}
\usepackage{booktabs}
\usepackage{graphicx}
\usepackage{amsmath}
\usepackage{amssymb}
\usepackage{multirow}
\usepackage{xcolor}
\usepackage{colortbl}
\usepackage{wrapfig}
\usepackage{capt-of}
\usepackage{float}
\usepackage[ruled,vlined]{algorithm2e}
\usepackage[section]{placeins}

\definecolor{DAMPBlue}{HTML}{5B7FA3}
\definecolor{DAMPBlueDark}{HTML}{3F668C}
\definecolor{DAMPBlueLight}{HTML}{E7EFF6}
\definecolor{ErrorTerracotta}{HTML}{C77C56}
\definecolor{ErrorTerracottaLight}{HTML}{F3E5DD}
\definecolor{PersistenceSage}{HTML}{6E9D8C}
\definecolor{PersistenceSageLight}{HTML}{E5EFEA}
\definecolor{GroupGray}{HTML}{ECEBE7}
\definecolor{MutedGray}{HTML}{7F858A}
\definecolor{LightGray}{HTML}{F4F3F0}

\newcommand{\khi}{K_{\mathrm{hi}}}
\newcommand{\method}{\textsc{Damp}}
\newcommand{\finding}[2]{%
  \par\smallskip
  \noindent\begingroup
  \setlength{\fboxsep}{5pt}%
  \colorbox{DAMPBlueLight}{%
    \parbox{\dimexpr\linewidth-2\fboxsep\relax}{%
      \normalsize\textbf{\textcolor{DAMPBlueDark}{Finding #1.}}~#2}}%
  \endgroup
  \par\smallskip
}

\title{DAMP: Decay-Aware Mixed-Precision\\Recurrent-State Quantization}

\author{Tao Zhang\textsuperscript{1,2}\thanks{Work done during internships at Meituan.}\quad 
Jianchao Tan\textsuperscript{2}\thanks{Corresponding author.} \quad 
Pingwei Sun\textsuperscript{2} \quad 
Yanqi Yu\textsuperscript{2,3} \quad 
Zunhai Su\textsuperscript{4} \\
\textbf{Zixu Jiang}\textsuperscript{2} \quad
\textbf{Yuchen Xie}\textsuperscript{2} \quad 
\textbf{Xunliang Cai}\textsuperscript{2} \quad 
\textbf{Ziqian Zeng}\textsuperscript{1}\footnotemark[2]\\
\textsuperscript{1}South China University of Technology \quad \textsuperscript{2}Meituan \quad \textsuperscript{3}East China Normal University \\
\textsuperscript{4}The University of Hong Kong \\
\texttt{tanjianchao02@meituan.com}\\
\texttt{zqzeng@scut.edu.cn}}

\iclrpreprintcopy
\ifdefined\ARXIVVERSION
  \iclrpreprintcopy
\fi
\begin{document}

\maketitle

\ificlrpreprint
  \fancyhead{}
  \fancyhead[L]{\small DAMP: Decay-Aware Mixed-Precision Recurrent-State Quantization}
  \renewcommand{\headrulewidth}{0.4pt}
\fi

\begin{abstract}
Complex reasoning and agentic applications increasingly rely on long-context inference, where growing KV caches increase both memory usage and decoding overhead. 
Hybrid models reduce these costs by combining Softmax Attention with Gated DeltaNet (GDN) or Kimi Delta Attention (KDA), which maintain fixed-size recurrent states.
These states are commonly stored in FP32 and consume substantial GPU memory, while their updates are limited by memory bandwidth.
Quantization can reduce both storage footprint and memory traffic, but we find that uniform INT8 and FP8 degrade complex reasoning accuracy, while INT4 and NVFP4 collapse it to near zero.
To our knowledge, this is the first study of post-training recurrent-state quantization for GDN and KDA.
Our analysis reveals that outliers in GDN and KDA states are concentrated in particular key channels and value dimensions.
Learned decay influences how much quantization error is retained.
We find that largely the same GDN heads and KDA key channels exhibit slow decay across tasks.
Based on these insights, we propose \method{}, which jointly considers quantization error and decay-based error retention to select high-risk key channels offline.
Under a fixed storage budget, it retains these channels in FP16 and stores the remainder in INT8.
We evaluate \method{} on Qwen3.6-35B, Kimi-Linear-48B and Kimi-K3 across six reasoning and code generation benchmarks.
At 9.9 bits per state value, \method{} maintains average accuracy close to FP32.
In SGLang, \method{} reduces recurrent-state storage by 69.1\%, accelerates the recurrent-state update kernel by up to $2.59\times$, and lowers full-model time per output token by up to 19.0\%.
\end{abstract}

\begin{figure}[H]
    \centering
    \includegraphics[width=0.95\linewidth]{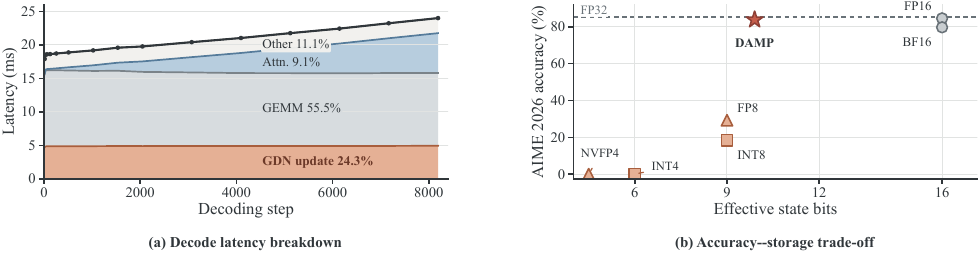}
    \caption{Recurrent-state update cost and accuracy--storage trade-off on Qwen3.6-35B.
    (a)~Decode latency breakdown at batch size 256.
    (b)~AIME 2026 accuracy versus effective state-storage bits.}
    \label{fig:bottleneck}
\end{figure}

\section{Introduction}
\label{sec:intro}

Complex reasoning and agentic applications increasingly rely on long contexts and extended sequences of reasoning and interaction \citep{kimiteam2025k15,kimiteam2026k3}. 
The growing KV cache, however, makes long-context inference memory-intensive.  
Recent models such as Qwen3.8 \citep{qwen2026qwen38, qwen2026design}, Qwen3.8-Flash-Next \citep{qwen2026qwen38flashnext}, Kimi-K3 \citep{kimiteam2026k3}, and GLM-5.3-Flash \citep{zai2026glm53flash} adopt hybrid architectures that interleave Gated DeltaNet (GDN) \citep{yang2024gateddeltanet} or Kimi Delta Attention (KDA) \citep{kimiteam2025kda} layers with full or sparse attention layers.

These linear-attention layers summarize past tokens in a fixed-size state matrix.
Serving systems commonly store these states in FP32.
For~example, serving Qwen3.6-35B \citep{qwen2026qwen36} with SGLang at batch size 256 requires an estimated 80~GB of GPU memory for FP32 recurrent states, including active states and checkpoints retained for prefix caching.
Recurrent states must also be read and updated at every decoding step.
These updates are limited by memory bandwidth and account for $24.3\%$ of decoding latency (Figure~\ref{fig:bottleneck}a).
State quantization therefore offers a direct way to reduce both the storage footprint and the memory traffic of recurrent updates.

However, recurrent states are difficult to quantize without degrading accuracy.
We find that uniform INT8 and FP8 quantization already degrades complex reasoning accuracy, while INT4 and NVFP4 reduce it to near zero (Figure~\ref{fig:bottleneck}b).
Recurrent states are updated and requantized at every decoding step.
In GDN and KDA, each update applies learned decay to the previous state and adds a delta-rule correction along the current key direction \citep{yang2024gateddeltanet,kimiteam2025kda}.
Quantization error therefore enters subsequent updates and propagates through the recurrence, as formalized in Section~\ref{sec:quantization_error_recurrence}.
Prior work has explored recurrent-state quantization in Mamba \citep{chiang2025quamba2,tianqi2025qmamba}.
To our knowledge, however, post-training quantization of recurrent states in GDN and KDA remains unexplored.

To understand this error, we examine the distribution of state magnitudes in GDN and KDA.
Their magnitudes vary markedly across key channels and value dimensions (Figure~\ref{fig:risk_structure}a--b).
After preprocessing along the value dimension, most quantization error is concentrated in a few key channels (Figure~\ref{fig:risk_structure}c).
These errors are subsequently affected by the recurrent updates: learned decay scales the previous state, while the delta-rule correction partially erases information along the current key direction.
Stronger decay suppresses more quantization error, while weaker decay allows more to persist.
Although this decay vary across decoding steps, largely the same GDN heads and KDA key channels exhibit slow decay across prompts and tasks (Figure~\ref{fig:risk_structure}d--f).

Motivated by these observations, we propose \method{}, which uses offline calibration to estimate each key channel's accumulated-error risk from its quantization error and decay-based error retention.
Under a fixed state-storage budget, it stores the highest-risk channels in higher precision and the remaining channels in the low-precision format.
We evaluate \method{} on Qwen3.6-35B-A3B, Kimi-Linear-48B-A3B-Instruct \citep{kimiteam2025kda}, and Kimi-K3 \citep{kimiteam2026k3}.
At 9.9 bits per state value, \method{} remains close to the FP32 baseline across the evaluated mathematical reasoning, general reasoning, and code generation benchmarks.
Relative to FP32-state inference, it reduces effective state storage by 69.1\%, accelerates the recurrent-update operator by up to $2.59\times$, and reduces full-model time per output token (TPOT) by up to 19.0\%.
Our contributions are summarized as follows:
\begin{itemize}
    \item To our knowledge, we are the first to study post-training quantization of recurrent states in GDN and KDA based language models.  
    We evaluate floating-point and integer formats and show that 8-bit quantization already causes substantial accuracy loss, with 4-bit formats degrading accuracy further.
    \item We find magnitude concentration along both dimensions of GDN and KDA states.
    After value-dimension preprocessing, quantization error is concentrated in a few key channels.
    We further find that largely the same GDN heads and KDA key channels exhibit slow decay across prompts and tasks.
    \method{} combines quantization error and decay-based error retention to allocate precision under a fixed storage budget.
    \item Across the evaluated benchmarks on Qwen3.6-35B-A3B, Kimi-Linear-48B-A3B-Instruct, and Kimi-K3, \method{} remains close to the FP32 baseline at 9.9 bits per state value.
    In SGLang, it reduces recurrent-state storage by 69.1\%, accelerates the recurrent-state update kernel by up to $2.59\times$, and lowers full-model TPOT by up to 19.0\%.
\end{itemize}

\section{Related Work}
\label{sec:related}

\paragraph{Post-training quantization of LLMs.}
Post-training quantization compresses model weights, activations, and KV caches to reduce inference costs.  
Recent work improves quantization accuracy through learned rotations and scaling transformations \citep{liu2025spinquant,hu2025ostquant} and mixed-precision allocation \citep{saxena2025resq}.  
KV-cache quantization reduces the memory cost of long-context inference \citep{liu2024kivi,zandieh2026turboquant,zhang2026mixkvq}.

\paragraph{Quantization and compression of recurrent states.} 
Prior work on quantizing linear-attention models has mainly focused on the Mamba family.  
Quamba and MambaQuant use activation-range control and rotation-based transformations to improve weight and activation quantization \citep{chiang2024quamba,xu2025mambaquant}.  
Quamba2 applies grouped quantization to cached recurrent states \citep{chiang2025quamba2}, while Q-Mamba uses separate scaling factors along the two state dimensions to reduce magnitude imbalance \citep{tianqi2025qmamba}.  
\method{} instead quantizes the recurrent states of GDN and KDA, using quantization error and decay-based error retention to select high-precision key channels offline.

\section{Recurrent-State Quantization in GDN and KDA}
\label{sec:background}

\subsection{GDN and KDA State Updates}

GDN and KDA store key--value associations in a recurrent state matrix.
For each head, the recurrent state is a matrix $\mS_t\in\mathbb{R}^{d_k\times d_v}$.
We denote the row corresponding to key channel $u$ by $\mS_{t,u}$.
At token step $t$, the query, key, and value vectors are $\vq_t,\vk_t\in\mathbb{R}^{d_k}$ and $\vv_t\in\mathbb{R}^{d_v}$.
Both architectures apply $\ell_2$ normalization to queries and keys within each head.
The query reads the updated state to produce
\begin{equation}
    \vo_t=\mS_t^{\top}\vq_t.
    \label{eq:state_readout}
\end{equation}

Both architectures apply learned decay to the previous state and then a delta-rule correction~\citep{yang2024gateddeltanet,kimiteam2025kda}:
\begin{equation}
    \mS_t
    =\bigl(\mI-\beta_t\vk_t\vk_t^{\top}\bigr)\mLambda_t\mS_{t-1}
    +\beta_t\vk_t\vv_t^{\top}.
    \label{eq:recurrence}
\end{equation}
Here, $\mLambda_t$ applies learned decay to the previous state, with diagonal entries $a_{t,u}$.
GDN uses a single decay factor $\alpha_t\in(0,1)$ shared across all key channels within a head, so $\mLambda_t=\alpha_t\mI$.
KDA instead applies channel-wise decay, with $\mLambda_t=\operatorname{diag}(\exp(\vg_t))$, where $\vg_t\in\mathbb{R}^{d_k}$.
The factor $\mI-\beta_t\vk_t\vk_t^{\top}$ partially erases the association with $\vk_t$ from the decayed state, while $\beta_t\vk_t\vv_t^{\top}$ writes the current value $\vv_t$ along the same key direction.
These two operations form the delta-rule correction, whose strength is controlled by $\beta_t\in(0,1)$.

We denote all operations applied to the previous state as the transition matrix, 
\begin{equation}
    \mA_t
    =\bigl(\mI-\beta_t\vk_t\vk_t^{\top}\bigr)\mLambda_t, 
    \label{eq:state_transition}
\end{equation}
Hence, Equation \ref{eq:recurrence} can be written as
\begin{equation}
    \mS_t
    = \mA_t \mS_{t-1}
    +\beta_t\vk_t\vv_t^{\top}.
    \label{eq:simplified_recurrence}
\end{equation}

\subsection{Quantized State Storage}
\label{sec:state_quantization}
Our quantization targets the stored recurrent state $\mS_t$ in Equation~\ref{eq:recurrence}.
During decoding, we reconstruct the previous state from quantized storage, perform the recurrent update, and quantize the updated state before writing it back to GPU memory.
We use symmetric $b$-bit integer quantization.
Let $\mathbf{x}$ denote a row of $\mS_t$.
The quantization scale is
\begin{equation}
    s=\frac{\max(|\mathbf{x}|)}{2^{b-1}-1}.
    \label{eq:integer_scale}
\end{equation}
The reconstructed values after quantization are
\begin{equation}
    Q(\mathbf{x})=s\,\operatorname{round}\!\left(\frac{\mathbf{x}}{s}\right).
    \label{eq:integer_quantizer}
\end{equation}
Specifications for all evaluated formats are provided in Appendix~\ref{app:quantizers}.

\subsection{Error Propagation under State Quantization}
\label{sec:quantization_error_recurrence}

The quantization operation at each step introduces error, and those errors propagate over time due to recurrent updates. 
Let $\mS_t^{\mathrm q}$ denote the state reconstructed from quantized storage at step $t$:
\begin{equation}
\mS_t^{\mathrm{q}}
=Q\!\left(
\mA_t\mS_{t-1}^{\mathrm{q}}+\beta_t\vk_t\vv_t^{\top}
\right).
\label{eq:quantized_recurrence}
\end{equation}

To measure accumulated quantization error up to step $t$, we define the
\textbf{accumulated state error} as the difference between the reconstructed state $\mS_t^{\mathrm q}$ and the unquantized state $\mS_t$ computed using Equation~\ref{eq:simplified_recurrence} without state quantization:
\begin{equation}
\Delta\mS_t=\mS_t^{\mathrm q}-\mS_t. 
\label{eq:state_deviation}
\end{equation}

To measure the error brought by the quantization operation at time step $t$, we define \textbf{local quantization error}, which is the difference between the output and input of $Q$ in Equation~\ref{eq:quantized_recurrence}:
\begin{equation}
\mR_t
=\mS_t^{\mathrm{q}}-\left(
\mA_t\mS_{t-1}^{\mathrm{q}}+\beta_t\vk_t\vv_t^{\top}
\right),
\label{eq:quantization_residual}
\end{equation}

Using Equation \ref{eq:simplified_recurrence} and 
$\mS_{t-1}=\mS_{t-1}^{\mathrm q} - \Delta\mS_{t-1}$ (derived from Equation \ref{eq:state_deviation}), we rewrite the accumulated state error as
\begin{align}
\Delta\mS_t
&= \mS_t^{\mathrm q} - \left( \mA_t \left(\mS_{t-1}^{\mathrm q} - \Delta\mS_{t-1} \right)
+\beta_t\vk_t\vv_t^{\top} \right)
\label{eq:deviation_expanded}\\
&= \mA_t \Delta\mS_{t-1} + \mS_t^{\mathrm q}-\left(
\mA_t\mS_{t-1}^{\mathrm q}+\beta_t\vk_t\vv_t^{\top}
\right)
\label{eq:deviation_second}\\
&=\mA_t\Delta\mS_{t-1}+\mR_t.
\label{eq:error_recurrence}
\end{align}

Unrolling Equation \ref{eq:error_recurrence} starting with $\Delta\mS_0=\bm{0}$, we have
\begin{equation}
\Delta\mS_t
=\mR_t
+\sum_{i=1}^{t-1}
\bigl(\mA_t\cdots\mA_{i+1}\bigr)\mR_i.
\label{eq:error_unrolled}
\end{equation}

From Equation~(\ref{eq:error_unrolled}), the accumulated state error depends not only on the local quantization errors introduced at each step but also on how subsequent state transitions transform these errors.
Both learned decay and the key-dependent factor in $\mA_t$ affect this propagation.

\section{Structure in Quantization Error and Decay}
\label{sec:findings}

We examine how the distribution of state magnitudes in GDN and KDA affects local quantization error, and whether learned decay provides a stable signal of error retention.

\begin{figure}[t]
\centering
\includegraphics[width=\linewidth]{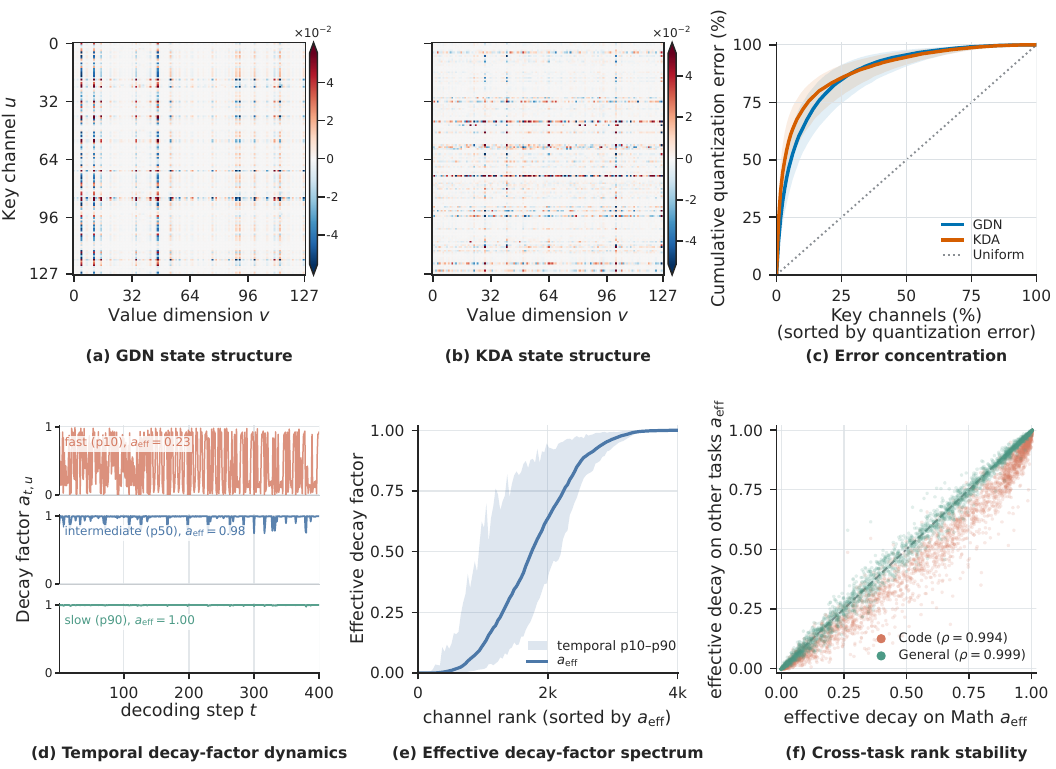}
\caption{
State structure, quantization error, and decay.
(a--b) Outliers across key channels and value dimensions in GDN and KDA.
(c) INT8 quantization error remains concentrated in a few key channels after value-dimension scaling and reordering.
(d--e) KDA decay varies across steps and channels.
(f) Channel rankings by effective decay factor remain stable across tasks.}
\label{fig:risk_structure}
\end{figure}

\subsection{State Structure and Error Concentration}
\label{sec:error_concentration}

Outliers in GDN and KDA states are concentrated in particular key channels and value dimensions (Figure~\ref{fig:risk_structure}a--b).
These outliers can lead to substantial quantization error in smaller-magnitude entries that share the same quantization scale.
We therefore quantize key channels independently, so that outliers in one channel do not affect the quantization scales of others.

Within each key channel, we divide state entries into groups and use the same quantization scale for all entries in a group.
Before quantization, we apply scaling and reordering (SR) along the value dimension.
We rescale each value dimension to reduce magnitude differences, then reorder the dimensions so that those with similar magnitudes share a quantization scale.
Applying SR to INT8 quantization increases the AIME 2026 accuracy of Qwen3.6-35B from 18.48\% to 63.98\%, still well below the FP32 baseline of 85.46\% (Table~\ref{tab:main}).

Figure~\ref{fig:risk_structure}c shows that this remaining error is highly concentrated across key channels.
The 12.5\% of key channels with the largest quantization errors account for approximately 69\% of the total squared error in GDN and 76\% in KDA.
This motivates storing a small subset of key channels in higher precision.

\finding{1}{
Outliers in GDN and KDA states are concentrated in particular key channels and value dimensions.
After value-dimension scaling and reordering, a small subset of key channels accounts for most of the quantization error.}

\subsection{Stable Structure in Learned Decay}
\label{sec:observation}
Equation~(\ref{eq:error_unrolled}) shows that local quantization errors propagate through subsequent state transitions.
Within these transitions, learned decay directly scales the error carried by the previous state.
We find substantial differences in decay strength across GDN heads and KDA key channels.
Figure~\ref{fig:risk_structure}d illustrates these differences using three KDA key channels with fast, intermediate, and slow decay.
The decay factor $a_{t,u}$ fluctuates substantially in the fast-decaying channel, whereas it remains close to one in the slow-decaying channel.

Since decay factors multiply across steps, we define the effective decay factor for key channel $u$ as their geometric mean, $a_{\mathrm{eff},u}=\exp(\mathbb E_t[\log a_{t,u}])$, where $\mathbb E_t$ averages over decoding steps.
An effective decay factor close to one indicates weak decay, which preserves much of the incoming state and its quantization error.
Figure~\ref{fig:risk_structure}e shows a broad range of effective decay factors across KDA key channels, from strong suppression to near-complete retention under decay.
GDN exhibits similarly broad variation across heads (Figure~\ref{fig:gdn_decay_structure}b).

We rank GDN heads and KDA key channels by their effective decay factors on Math, Code, and General.
Although decay factors vary across decoding steps, largely the same heads and key channels exhibit slow decay across tasks.
For KDA, the Spearman rank correlations with Math are 0.994 for Code and 0.999 for General (Figure~\ref{fig:risk_structure}f).
GDN shows similarly stable rankings at the head level (Figure~\ref{fig:gdn_decay_structure}c).
This cross-task stability supports offline estimation of decay-based error retention.

\finding{2}{
Decay strength varies substantially across GDN heads and KDA key channels.
Although decay factors vary across decoding steps, largely the same heads and key channels exhibit slow decay across tasks.}

\section{DAMP: Decay-Aware Mixed-Precision State Quantization}
\label{sec:method}

\begin{figure}[t!]
    \centering
    \includegraphics[width=\linewidth]{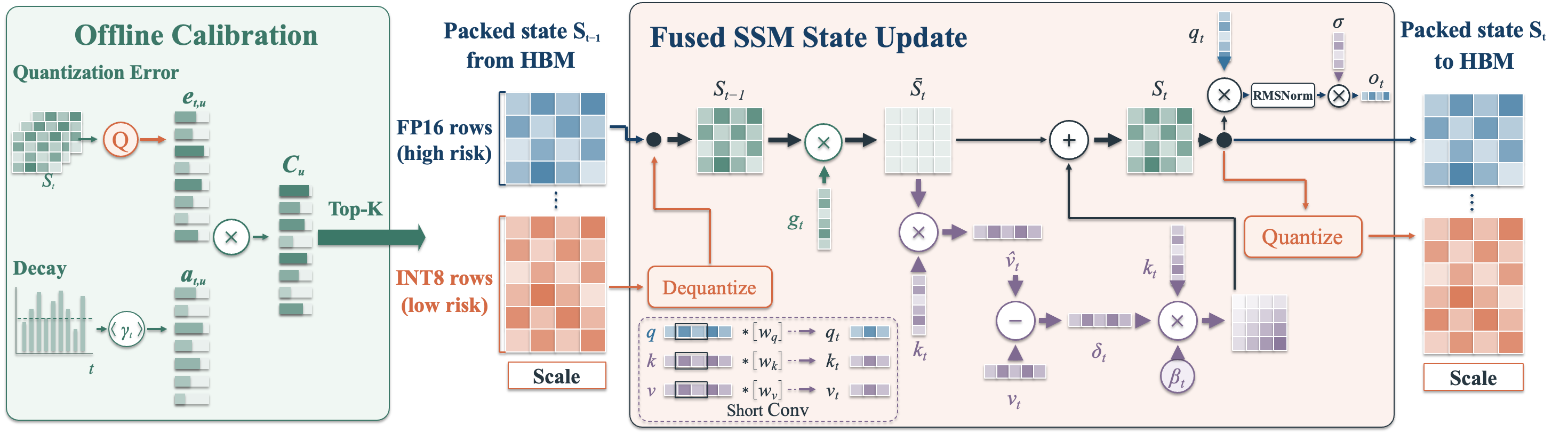}
    \caption{
    Overview of \method{}.  
    Offline calibration selects high-precision key channels.
    A fused kernel updates the packed FP16/INT8 state, combining low-precision reconstruction, the recurrence, and requantization.}
    \label{fig:damp_workflow}
\end{figure}

\subsection{Budgeted Key-Channel Allocation}
\label{sec:budgeted_allocation}

\paragraph{Quantization-error measurement.}
Our goal is to protect key channels whose quantization errors have the largest estimated cumulative impact under a limited high-precision budget.
For each layer and head, we estimate local quantization error (Section~\ref{sec:quantization_error_recurrence}) on sampled calibration states using the quantizer in Section~\ref{sec:error_concentration}.
For key channel $u$, its squared magnitude is
\begin{equation}
    e_{t,u}
    =\bigl\|[Q(\mS_t)-\mS_t]_u\bigr\|_2^2.
    \label{eq:error_energy}
\end{equation}

\paragraph{Accumulated-error risk.}
We estimate the contribution of each local quantization error to subsequent states under learned decay.
Using the observed decay factors, the estimated squared-error contribution after $j$ updates is $e_{t,u}\prod_{r=1}^{j}a_{t+r,u}^{2}$.
We sum the initial squared error and these decay-weighted contributions over the next $H$ updates, then average over calibration samples to define the accumulated-error risk score for key channel $u$:
\begin{equation}
    C_u
    =\mathbb E_{\mathrm{cal}}\!\left[
        e_{t,u}\left(
            1+\sum_{j=1}^{H}\prod_{r=1}^{j}a_{t+r,u}^{2}
        \right)
      \right].
    \label{eq:cumulative_decay_error}
\end{equation}
Here, $\mathbb E_{\mathrm{cal}}$ averages over sampled calibration states with $H$ subsequent updates available within the same trajectory.
The score favors key channels with large local quantization errors and weak subsequent decay.

\paragraph{Budgeted selection.}
Let $\khi$ denote the number of key channels stored in high precision per head.
We use the same count across layers and heads to maintain a uniform memory layout and simplify fused recurrent updates.
For each layer $\ell$ and head $h$, we select the $\khi$ channels with the highest scores $C_{\ell,h,u}$, and denote their indices by $\mathcal H_{\ell,h}$.
For a fixed channel count, this selection minimizes the total risk score of the channels left in low precision.
The selected indices are determined independently for each layer and head and remain fixed throughout inference.

\subsection{Packed Layout and Fused State Update}

We reorder the key channels so that the selected high-precision channels occupy the first $\khi$ rows of the state matrix, followed by the low-precision channels.
Before inference, we rearrange the rows of the query and key projection weights to produce their outputs in this order.
The value-dimension reordering used by SR is implemented similarly: we rearrange the rows of the value projection weights and the corresponding columns of the output projection weights.
Convolution, gating, and normalization parameters are reordered to match.
The model then produces recurrent inputs in the required channel order, eliminating explicit reordering during decoding.

We choose FP16 for the high-precision channels, as FP16 state achieves accuracy close to FP32 (Table~\ref{tab:main}).
The packed state consists of contiguous FP16 and INT8 regions.
At each decoding step, a single CUDA kernel fuses state reconstruction, the recurrent update, and requantization before writing back the packed state (Figure~\ref{fig:damp_workflow}).
Implementation details are provided in Appendix~\ref{app:efficient_implementation}.

\newcommand{\accse}[2]{#1}
\begin{table}[t]
\caption{
Main results on Qwen3.6-35B-A3B and Kimi-Linear-48B-A3B-Instruct. 
``Avg.\ bits'' reports effective state-storage cost per element, including scales and zero points.  
SR denotes scaling and reordering along the value dimension.
\method{} results are highlighted in \textbf{bold}.
}
\label{tab:main}
\centering
\setlength{\tabcolsep}{3.5pt}
\resizebox{\textwidth}{!}{
\begin{tabular}{lccccccc}
\toprule
 & & \multicolumn{3}{c}{\textbf{Math Reasoning}} & \multicolumn{2}{c}{\textbf{General Reasoning}} & \textbf{Code} \\
\cmidrule(lr){3-5}\cmidrule(lr){6-7}\cmidrule(lr){8-8}
\textbf{Method} & \textbf{Avg.\ bits} & \textbf{AIME 2026} & \textbf{HMMT} & \textbf{IMO-Ans} & \textbf{GPQA-D} & \textbf{MMLU-Pro} & \textbf{LCB-v6} \\
\midrule
\rowcolor{GroupGray}\multicolumn{8}{l}{\emph{Qwen3.6-35B-A3B}} \\
\addlinespace[2pt]
FP32            & 32   & \accse{85.46}{0.38} & \accse{57.57}{0.61} & \accse{50.29}{0.28} & \accse{81.97}{0.18} & \accse{84.66}{0.02} & \accse{86.95}{0.28} \\
FP16            & 16   & \accse{84.58}{0.39} & \accse{55.82}{0.62} & \accse{49.42}{0.28} & \accse{82.13}{0.23} & \accse{84.61}{0.02} & \accse{86.80}{0.29} \\
BF16            & 16   & \accse{79.71}{0.43} & \accse{54.97}{0.59} & \accse{43.71}{0.29} & \accse{81.66}{0.23} & \accse{84.65}{0.02} & \accse{84.10}{0.28} \\
\addlinespace[2pt]
FP8 (E4M3)      & 9.0  & \accse{29.27}{0.92} & \accse{14.06}{0.38} & \accse{10.23}{0.20} & \accse{76.26}{0.36} & \accse{77.66}{0.06} & \accse{15.02}{0.34} \\
INT8            & 9.0  & \accse{18.48}{0.39} & \accse{20.12}{0.47} & \accse{16.38}{0.27} & \accse{79.48}{0.29} & \accse{83.88}{0.03} & \accse{32.23}{0.28} \\
INT8+SR         & 9.0  & \accse{63.98}{0.58} & \accse{50.09}{0.83} & \accse{39.31}{0.54} & \accse{81.89}{0.24} & \accse{84.59}{0.03} & \accse{64.51}{0.28} \\
\addlinespace[2pt]
NVFP4           & 4.5  & \accse{0.03}{0.01} & \accse{0.05}{0.01} & \accse{0.98}{0.03} & \accse{16.57}{0.55} & \accse{9.31}{0.05} & \accse{1.02}{0.02} \\
INT4            & 6.0  & \accse{0.07}{0.01} & \accse{0.14}{0.02} & \accse{1.03}{0.03} & \accse{28.69}{0.61} & \accse{32.19}{0.10} & \accse{0.85}{0.01} \\
INT4+SR         & 6.0  & \accse{0.03}{0.01} & \accse{0.09}{0.01} & \accse{1.02}{0.04} & \accse{24.59}{0.69} & \accse{13.28}{0.08} & \accse{0.02}{0.00} \\
\addlinespace[2pt]
\rowcolor{DAMPBlueLight}
\textbf{\method{} INT8}  & 9.9  & \textbf{83.82} & \textbf{55.11} & \textbf{48.89} & \textbf{82.05} & \textbf{84.57} & \textbf{85.67} \\
\midrule
\rowcolor{GroupGray}\multicolumn{8}{l}{\emph{Kimi-Linear-48B-A3B-Instruct}} \\
\addlinespace[2pt]
FP32            & 32   & \accse{62.34}{0.55} & \accse{38.02}{0.63} & \accse{28.04}{0.28} & 64.17 & \accse{68.48}{0.07} & \accse{61.17}{0.34} \\
FP16            & 16   & \accse{63.07}{0.56} & 37.83 & \accse{28.10}{0.28} & \accse{64.74}{0.58} & \accse{68.42}{0.07} & \accse{60.58}{0.35} \\
BF16            & 16   & \accse{58.56}{0.56} & \accse{32.77}{0.60} & \accse{25.37}{0.26} & \accse{63.48}{0.57} & \accse{68.51}{0.07} & \accse{60.79}{0.37} \\
\addlinespace[2pt]
FP8 (E4M3)      & 9.0  & \accse{29.47}{0.62} & \accse{13.16}{0.26} & \accse{11.20}{0.19} & \accse{54.26}{0.71} & \accse{56.90}{0.11} & \accse{22.16}{0.46} \\
INT8            & 9.0  & \accse{49.90}{0.54} & \accse{27.08}{0.50} & \accse{22.21}{0.24} & \accse{63.22}{0.56} & \accse{68.37}{0.07} & \accse{57.97}{0.36} \\
INT8+SR         & 9.0  & \accse{58.07}{0.54} & \accse{32.05}{0.83} & \accse{24.67}{0.45} & \accse{62.61}{0.59} & \accse{68.47}{0.07} & \accse{59.49}{0.34} \\
\addlinespace[2pt]
NVFP4           & 4.5  & \accse{3.38}{0.10} & \accse{1.47}{0.10} & \accse{3.31}{0.08} & \accse{46.75}{0.72} & \accse{58.47}{0.09} & \accse{6.04}{0.07}  \\
INT4            & 6.0  & \accse{7.50}{0.13} & \accse{2.46}{0.12} & \accse{6.53}{0.13} & \accse{46.65}{0.70} & \accse{61.00}{0.08} & \accse{0.79}{0.02}  \\
INT4+SR         & 6.0  & \accse{0.52}{0.04} & \accse{0.47}{0.04} & \accse{2.42}{0.06} & \accse{40.02}{0.76} & \accse{55.68}{0.08} & \accse{2.49}{0.05} \\
\addlinespace[2pt]
\rowcolor{DAMPBlueLight}
\textbf{\method{} INT8}  & 9.9  & \textbf{62.94} & \textbf{38.54} & \textbf{27.75} & \textbf{64.56} & \textbf{68.64} & \textbf{61.02} \\
\bottomrule
\end{tabular}
}
\end{table}

\section{Experiments}
\label{sec:experiments}

We evaluate \method{} in three stages.  We first report downstream accuracy at the target storage budget, then measure recurrent-update and end-to-end serving efficiency, and finally ablate the key-channel selector and precision budget.

\subsection{Experimental Setup}
\label{sec:experimental_setup}

\paragraph{Models.}
We evaluate Qwen3.6-35B-A3B \citep{qwen2026qwen36} and Kimi-Linear-48B-A3B-Instruct \citep{kimiteam2025kda}, two hybrid MoE models with approximately 3B activated parameters.
Qwen3.6 uses 30 GDN and 10 full-attention layers, while Kimi-Linear uses 20 KDA and 7 MLA layers.
We also evaluate Kimi-K3 \citep{kimiteam2026k3}, a 2.8T-parameter hybrid MoE model with 104B activated parameters, 69 KDA layers, and 24 gated MLA layers.

\paragraph{Benchmarks.}
Mathematical reasoning uses AIME 2026 Parts~I and II and HMMT February 2026 \citep{dekoninck2026matharena}, as well as IMO-AnswerBench \citep{luong-etal-2025-towards}.  
General reasoning uses GPQA-Diamond and MMLU-Pro \citep{rein2024gpqa,wang2024mmlu}; code generation uses LiveCodeBench-v6 \citep{livecodebench}.

\paragraph{Baselines.}
We implement all methods in SGLang \citep{zheng2024sglang}, using its default FP32 recurrent-state storage as the reference.
Baselines include FP16 and BF16 storage, FP8 and NVFP4 quantization, and INT8/INT4 with and without SR.
Quantizer specifications and generation settings are provided in Appendices~\ref{app:quantizers} and~\ref{app:evaluation_settings}, respectively.

\paragraph{Calibration.}
We calibrate each model using 36 prompts spanning mathematical reasoning, code generation, and general tasks.
Each input contains 1,024 tokens, and the model generates 4,096 tokens using FP32 recurrent-state storage.
We compute the channel scores in Equation~(\ref{eq:cumulative_decay_error}) from quantization errors measured on sampled states and the observed decay factors over the next \mbox{$H=256$} updates after each sampled state.
In the main experiments, \method{} stores $\khi=16$ key channels per head in FP16 and the remainder in INT8+SR, requiring 9.9 bits per state value.
Data sources, splits, and state-sampling details are provided in Appendix~\ref{app:calibration}.

\begin{figure}[t!]
\centering
\includegraphics[width=\linewidth]{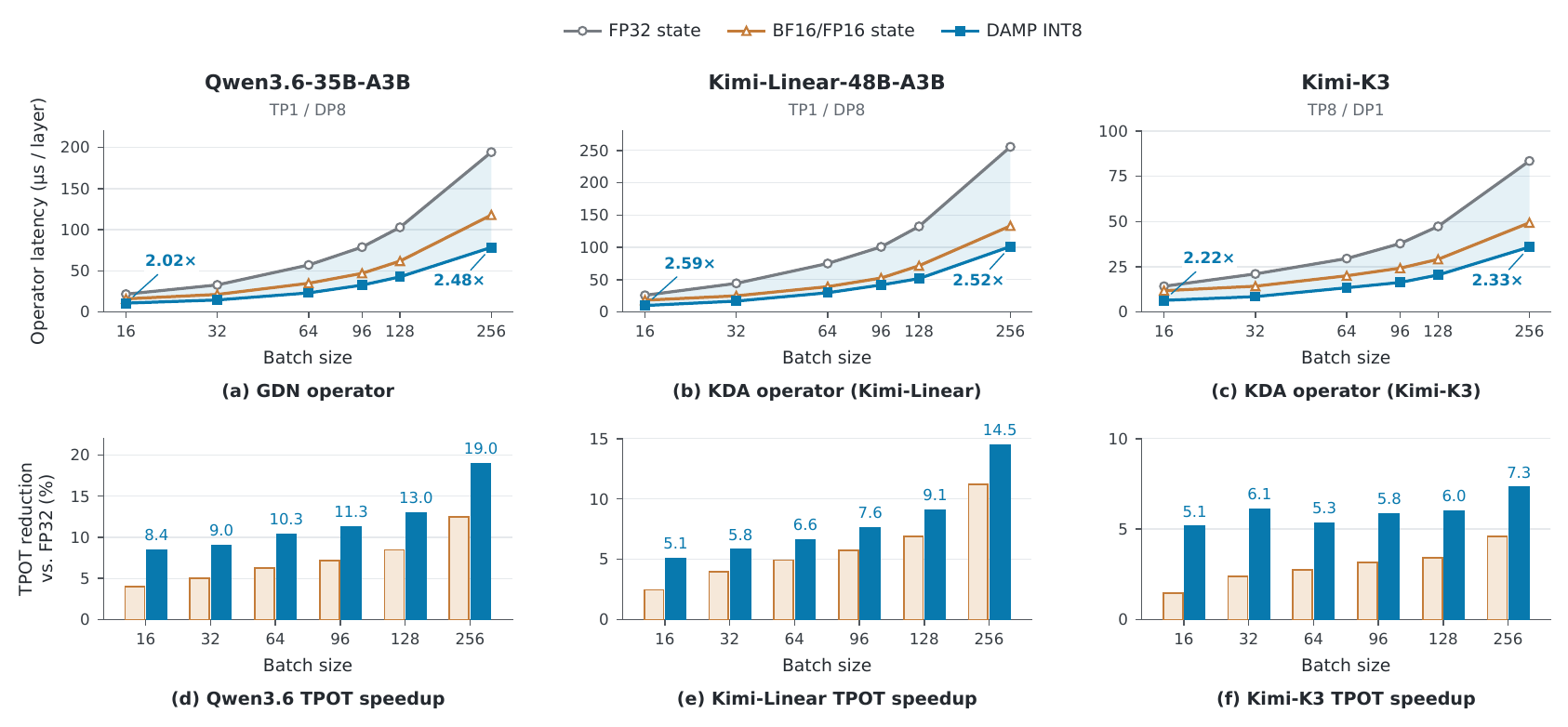}
\caption{
Decoding efficiency of \method{}-INT8, FP32, and BF16/FP16 state storage.
(a--c) Recurrent-update latency per layer; 
(d--f) TPOT reduction relative to FP32.
Columns show Qwen3.6-35B-A3B, Kimi-Linear-48B-A3B, and Kimi-K3, respectively.}
\label{fig:efficiency_main}
\vspace{-1.0em}
\end{figure}

\subsection{Main Results}
\label{sec:main_results}

\noindent
\begin{minipage}[t]{0.59\linewidth}
\vspace{0pt}
Table~\ref{tab:main} compares \method{} with full-precision references and uniform state quantization.
With only 0.9 additional bits per state value, \method{} recovers much of the accuracy lost under uniform INT8+SR.
For example, \method{} improves AIME 2026 accuracy on Qwen3.6-35B by 19.84 percentage points over uniform INT8+SR.
The additional evaluation on Kimi-K3 extends these results to a larger model: at 9.9 bits per state value, \method{} achieves accuracy close to FP32 on both AIME 2026 and HMMT (Table~\ref{tab:kimi_k3_math}).
\end{minipage}\hfill
\begin{minipage}[t]{0.37\linewidth}
\vspace{0pt}
\setlength{\abovecaptionskip}{0pt}
\setlength{\belowcaptionskip}{4pt}
\captionof{table}{Kimi-K3 accuracy (\%).}
\label{tab:kimi_k3_math}
\centering
\scriptsize
\setlength{\tabcolsep}{2.2pt}
\renewcommand{\arraystretch}{1.02}
\begin{tabular}{@{}lccc@{}}
\toprule
\textbf{Method} & \textbf{Avg.\ bits} & \textbf{AIME 2026} & \textbf{HMMT} \\
\midrule
FP32           & 32  & 96.88 & 97.92 \\
BF16           & 16  & 97.29 & 96.78 \\
FP8            & 9.0 & 59.83 & 40.91 \\
INT8           & 9.0 & 89.17 & 82.58 \\
INT8+SR        & 9.0 & 93.96 & 91.67 \\
INT4           & 6.0 &  4.38 &  7.58 \\
\rowcolor{DAMPBlueLight}
\textbf{\method{} INT8} & \textbf{9.9} & \textbf{97.50} & \textbf{97.35} \\
\bottomrule
\end{tabular}
\end{minipage}
\par\smallskip

FP16 maintains accuracy close to FP32 on both Qwen3.6 and Kimi-Linear, while outperforming BF16 on average at the same 16-bit storage cost.
This contrast suggests that these recurrent states benefit more from FP16's finer numerical resolution than from BF16's wider exponent range.
Kimi-K3 is less sensitive to BF16 state storage, retaining accuracy close to FP32 on both AIME 2026 and HMMT.
We speculate that quantization-aware training for BF16 recurrent-state storage may contribute to this robustness.
With INT8+SR, accuracy losses vary substantially across tasks.
On Qwen3.6-35B, accuracy remains within 0.1 percentage points of FP32 on GPQA-Diamond and MMLU-Pro, but drops by more than 20 points on AIME 2026 and LiveCodeBench-v6.
INT4 and NVFP4 cause severe accuracy degradation in mathematical reasoning and code generation, making the evaluated 4-bit configurations impractical for these tasks.

\subsection{Efficiency}
\label{sec:efficiency_results}

We compare \method{} with FP32 and 16-bit (BF16/FP16) state storage in SGLang at decode batch sizes from 16 to 256.
Qwen3.6 and Kimi-Linear use tensor parallelism (TP) of 1 and data parallelism (DP) of 8, while Kimi-K3 uses TP8/DP1.
Requests are sampled from the ShareGPT-V3 conversation dataset\footnote{\url{https://huggingface.co/datasets/anon8231489123/ShareGPT_Vicuna_unfiltered}}, with input and output lengths fixed at 256 and 64 tokens, respectively.
We report per-layer recurrent-update latency, including state loads, dequantization, quantization, and writeback, together with full-model time per output token (TPOT).

Figure~\ref{fig:efficiency_main} shows that \method{} reduces recurrent-update latency across all three models, outperforming both FP32 and 16-bit state storage and achieving up to a $2.59\times$ speedup over FP32.
These gains translate into faster full-model decoding: at batch size 256, \method{} reduces TPOT relative to FP32 by 19.0\% on Qwen3.6, 14.5\% on Kimi-Linear, and 7.3\% on Kimi-K3.
The smaller TPOT reduction on Kimi-K3 may partly reflect the inter-device communication overhead of its TP8 configuration.

\noindent
\begin{minipage}[t]{0.59\linewidth}
\vspace{0pt}
\paragraph{Multi-turn serving.}
Beyond decoding latency, we evaluate time to first token (TTFT) and prefix-cache reuse on Kimi-K3 under a fixed total memory budget for recurrent states and the KV cache.
The workload consists of 256 four-turn conversations from WildChat-1M \citep{zhao2024wildchat}.
Compared with FP32 and BF16, \method{} reduces mean TTFT by 20.7\% and 14.5\%, respectively, while achieving higher prefix-cache hit rates (Table~\ref{tab:k3_multiturn_ttft}).
Workload details are provided in Appendix~\ref{app:k3_multiturn_ttft}.
\end{minipage}\hfill
\begin{minipage}[t]{0.37\linewidth}
\vspace{0pt}
\setlength{\abovecaptionskip}{0pt}
\setlength{\belowcaptionskip}{4pt}
\captionof{table}{Multi-turn TTFT and cache hit rates on Kimi-K3.}
\label{tab:k3_multiturn_ttft}
\centering
\scriptsize
\setlength{\tabcolsep}{2.2pt}
\renewcommand{\arraystretch}{1.02}
\begin{tabular}{@{}lccc@{}}
\toprule
\multirow{2}{*}{\textbf{State storage}} &
\multicolumn{2}{c}{\textbf{TTFT (s)} $\downarrow$} &
\multirow{2}{*}{\shortstack{\textbf{Mean cache}\\\textbf{hit rate (\%)} $\uparrow$}} \\
\cmidrule(lr){2-3}
& Mean & P95 & \\
\midrule
FP32 & 38.427 & 72.471 & 47.20 \\
BF16 & 35.646 & 69.500 & 56.96 \\
\rowcolor{DAMPBlueLight}
\textbf{\method{} INT8} & \textbf{30.464} & \textbf{61.559} & \textbf{65.29} \\
\bottomrule
\end{tabular}
\end{minipage}
\par\smallskip

\subsection{Ablation Studies}
\label{sec:ablations}

\noindent
\begin{minipage}[t]{0.63\linewidth}
\vspace{0pt}
\paragraph{Key-channel selector.}
We compare the criteria used to select high-precision key channels under the same INT8+SR quantizer and storage budget (Table~\ref{tab:score_ablation_main}).
Random selects channels uniformly, while State energy ranks them by their mean squared norm.
Error-only selection ranks channels by their average local quantization error, while retention-only selection ranks them by their average decay-based error retention weight.
\method{} achieves the highest accuracy on all three benchmarks, which supports incorporating error retention into channel selection rather than relying on local quantization error alone.
Additional GDN results are provided in Appendix~\ref{app:gdn_ablations}.
\par\vspace{6pt}
\paragraph{Precision budget.}
Figure~\ref{fig:precision_budget} shows AIME 2026 accuracy as we vary the number of key channels retained in FP16.
Increasing the FP16 budget generally improves accuracy, but provides limited additional gains beyond $\khi=16$.
These results suggest that protecting a small subset of key channels can recover most of the accuracy lost to INT8+SR quantization.
We therefore use $\khi=16$ in the main experiments.
\end{minipage}\hfill
\begin{minipage}[t]{0.33\linewidth}
\vspace{0pt}
\setlength{\parskip}{0pt}
\setlength{\abovecaptionskip}{0pt}
\setlength{\belowcaptionskip}{4pt}
\captionof{table}{Matched-budget KDA selector ablation at 9.9 bits per state value.}
\label{tab:score_ablation_main}
\centering
\begingroup
\scriptsize
\setlength{\tabcolsep}{3pt}
\renewcommand{\arraystretch}{1.08}
\begin{tabular}{@{}lccc@{}}
\toprule
\textbf{Selector} & \textbf{AIME} & \textbf{GPQA} &
\textbf{LCB} \\
\midrule
\rowcolor{GroupGray}\multicolumn{4}{l}{\emph{Kimi-Linear (KDA)}} \\
Random                    & 59.06         & 62.12         & 58.68         \\
State eng.                & 60.36         & 63.38         & 59.67         \\
Error-only                & 61.40         & 63.85         & 60.34         \\
Retention-only            & 61.22         & 63.69         & 60.69         \\
\rowcolor{DAMPBlueLight}
\textbf{\method{} ($C$)}
                          & \textbf{62.94} & \textbf{64.56} & \textbf{61.02} \\
\bottomrule
\end{tabular}\par
\endgroup
\vspace{8pt}
\includegraphics[width=0.9\linewidth]{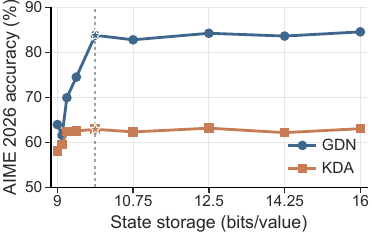}
\setlength{\abovecaptionskip}{3pt}
\setlength{\belowcaptionskip}{0pt}
\captionof{figure}{AIME 2026 \mbox{accuracy} versus bits per state value.}
\label{fig:precision_budget}
\end{minipage}
\par\vspace{-1pt}

\section{Conclusion}

We introduced \method{}, a post-training method for compressing the recurrent states of GDN- and KDA-based language models.  
It ranks key channels using quantization error and decay-based error retention estimated during calibration, then realizes the allocation as a static mixed-precision layout.  
At 9.9 bits per state value, \method{} maintains average accuracy close to FP32 across the three evaluated checkpoints. 
Relative to FP32-state inference, it reduces state storage by 69.1\%, accelerates the recurrent-state update kernel by up to $2.59\times$, and lowers full-model TPOT by up to 19.0\%.
\subsection*{AI use statement}

We used AI for language editing, code debugging, and figure and table
prototyping.  The authors reviewed and revised all AI-assisted text.  All
measurements reported in this paper were produced by the evaluation procedures
described in the paper.  The authors take full responsibility for the final
content, claims, and research artifacts.

\subsection*{Ethics statement}

This work studies an inference-time compression method using publicly released
model checkpoints, calibration corpora, and established evaluation benchmarks.
It does not involve human participants or the collection of private data.  The
method changes the storage and execution of recurrent states rather than the
intended capabilities of the underlying models; consequently, deployments
retain the limitations and potential risks of those models and their training
data.

\subsection*{Reproducibility statement}

Section~\ref{sec:experimental_setup} describes the evaluated models, benchmarks, baselines, and calibration setup.
Appendices~\ref{app:quantizers}--\ref{app:efficient_implementation} describe the quantization formats, calibration procedure, key-channel selection algorithm, and implementation.
Generation settings and the number of generations per problem are provided in Appendix~\ref{app:evaluation_settings}.
Section~\ref{sec:efficiency_results} and Appendix~\ref{app:k3_multiturn_ttft} describe the serving evaluation setup.

\bibliography{iclr2027_conference}
\bibliographystyle{iclr2027_conference}
\newpage
\appendix
\renewcommand{\theHfigure}{appendix.\arabic{figure}}
\renewcommand{\theHtable}{appendix.\arabic{table}}
\renewcommand{\theHalgocf}{appendix.\arabic{algocf}}
\section{Additional Method Details}
\label{app:quantization_details}

\subsection{Storage Quantizers}
\label{app:quantizers}

All groupwise quantizers partition the value dimension of each key channel into contiguous blocks.  
Group quantization parameters are recomputed whenever the state is written.

\paragraph{Integer formats.}
INT8 uses the symmetric mapping in Equation~\ref{eq:integer_quantizer} with groups of 32 values.  Each group stores one FP32 scale and no zero point.
INT4 uses asymmetric quantization with the same group size, storing one FP32 scale $s$ and one FP32 zero point $z$ per group.

\paragraph{Value scaling and reordering.}
We rescale each value dimension to reduce magnitude differences.
The scaling factors are calibrated separately for each layer and head.
We then sort the value dimensions by their RMS magnitudes after scaling, placing dimensions with similar magnitudes next to each other.
Each low-precision key channel is quantized in groups of 32 consecutive entries, with one quantization scale per group.
The scaling factors and value order are selected during offline calibration and remain fixed during inference.
In contrast, group quantization parameters are recomputed at each state write.

\paragraph{Floating-point formats.}
FP8 uses E4M3 without SR, with one FP32 scale per group of 32 state values and no zero point.
The scale is computed as the group's maximum absolute value divided by 448.
Values are divided by this scale before conversion to E4M3 and multiplied by it during reconstruction.
NVFP4 uses blocks of 16 E2M1 values, one E4M3 scale per block, and one FP32 global scale.

\subsection{Calibration Protocol}
\label{app:calibration}

\paragraph{Data and splits.}
All three models use 36 prompts: 12 from the MATH training split~\citep{hendrycks2021math}, 12 from the APPS training split~\citep{hendrycks2021apps}, and 12 from the ARC-Challenge training split~\citep{clark2018arc}. 
We divide these prompts into disjoint fitting, parameter-selection, and validation splits of 12 each.
Every split contains four mathematical, four code, and four general prompts.

\paragraph{Reference trajectories.}
Each input contains 1,024 tokens, including the chat template and special tokens.
The model generates 4,096 tokens with FP32 recurrent-state storage; early termination is suppressed to maintain the fixed output length.
We record states and decay factors over two windows per sequence, starting at decoding-update positions 0 and 2,048 and covering 1,024 updates each.
The states after the 256th, 512th, and 768th updates within each window serve as scoring anchors.
For each anchor, the following $H=256$ updates provide a complete decay sequence for Equation~\ref{eq:cumulative_decay_error}.
This gives six anchors per sequence and 72 per split.

\paragraph{Scaling and channel selection.}
For each layer and head, we use state RMS statistics from the fitting split to construct candidate value-scaling factors and value orders (Appendix~\ref{app:quantizers}).
For each candidate $c$, we simulate INT8 quantization at the fitting anchors and restore the original value scales and channel order before measuring $e_{t,u}$ in Equation~\ref{eq:error_energy}.
We compute $C_u$ using the next 256 observed decay factors as in Equation~\ref{eq:cumulative_decay_error} and select the $\khi$ highest-scoring key channels to form the FP16 set $\mathcal H_c$.

We then freeze $\mathcal H_c$ and evaluate the complete mixed-precision configuration on the parameter-selection split:
\begin{equation}
    J(c)=\mathbb E_{\mathrm{tune}}\!\left[
        \sum_{u\notin\mathcal H_c} W_{t,u}\,e^{(8,c)}_{t,u}
        +\sum_{u\in\mathcal H_c} W_{t,u}\,e^{(16,c)}_{t,u}
    \right].
    \label{eq:sr_candidate_objective}
\end{equation}
Here, $e^{(8,c)}_{t,u}$ and $e^{(16,c)}_{t,u}$ are the squared errors after INT8 quantization and FP16 storage, respectively, both measured in the original state coordinates. 
$W_{t,u}$ is the decay-based error retention weight in Equation~\ref{eq:cumulative_decay_error}, including the initial contribution with weight one, and $\mathbb E_{\mathrm{tune}}$ averages over the parameter-selection split.
All calibration averages give equal weight to anchors within a window, windows within a sequence, sequences within a domain, and the three domains.
We choose the candidate with the smallest $J(c)$ and retain its fitted protected set and key permutation (Algorithm~\ref{alg:calib}).
The independent validation split evaluates the frozen configuration.

\begin{algorithm}[t]
\caption{Offline calibration and layout selection}
\label{alg:calib}
\KwIn{fitting and parameter-selection trajectories with scoring anchors and
per-update decay factors; $H=256$; high-precision key-channel count $\khi$}
\KwOut{SR configurations, protected key-channel sets, and key permutations}
\ForEach{recurrent layer $\ell$ and head $h$}{
  Construct SR candidates from state RMS statistics on the fitting split\;
  \ForEach{candidate $c$}{
    Compute $C_{\ell,h,:}^{(c)}$ on fitting anchors using
    Equations~\ref{eq:error_energy} and~\ref{eq:cumulative_decay_error}\;
    $\mathcal H_c\leftarrow\operatorname{TopK}(C_{\ell,h,:}^{(c)},\khi)$\;
    Evaluate $J(c)$ on the parameter-selection split with $\mathcal H_c$
    fixed, using Equation~\ref{eq:sr_candidate_objective}\;
  }
  $c^*_{\ell,h}\leftarrow\operatorname*{arg\,min}_{c}J(c)$\;
  $\mathcal H_{\ell,h}\leftarrow\mathcal H_{c^*_{\ell,h}}$\;
  Construct $\pi_{\ell,h}$ so that the selected channels occupy the first
  $\khi$ state rows, preserving channel order within each precision region\;
}
\Return{$\{c^*_{\ell,h},\mathcal H_{\ell,h},\pi_{\ell,h}\}_{\ell,h}$}\;
\end{algorithm}

\paragraph{Calibration cost.}
Calibration takes approximately 7 minutes for Qwen3.6-35B, 9 minutes for Kimi-Linear-48B, and 78.75 minutes for Kimi-K3.

\subsection{State Scaling and Reordering}
\label{app:state_transform}

Algorithm~\ref{alg:calib} constructs a key-channel permutation that places the selected high-precision channels in the first $\khi$ rows of the state matrix.
Along the value dimension, SR applies the scaling and reordering described in Appendix~\ref{app:quantizers}.
To update the state in this representation, the recurrent inputs must use the same channel orders and value scales.

For one head, let $\mathbf P_K$ and $\mathbf P_V$ denote the fixed permutation matrices for key channels and value dimensions, respectively.
Let $\mathbf D_c=\operatorname{diag}(c_1,\ldots,c_{d_v})$, with $c_v>0$, contain the value-scaling factors.
The transformed state $\mT_t$ and recurrent inputs are
\begin{equation}
\begin{aligned}
    \mT_t&=\mathbf P_K\mS_t\mathbf D_c^{-1}\mathbf P_V^{\top},\\
    \vq'_t&=\mathbf P_K\vq_t,\qquad
    \vk'_t=\mathbf P_K\vk_t,\qquad
    \vv'_t=\mathbf P_V\mathbf D_c^{-1}\vv_t.
\end{aligned}
\label{eq:permuted_state_coordinates}
\end{equation}
The decay factors follow the same key-channel order, giving
$\mLambda'_t=\mathbf P_K\mLambda_t\mathbf P_K^{\top}$ and
$\mA'_t=\mathbf P_K\mA_t\mathbf P_K^{\top}$.
The scaling factors in the reordered value coordinates form
$\mathbf D'_c=\mathbf P_V\mathbf D_c\mathbf P_V^{\top}$.
Substituting these transformations into Equations~\ref{eq:state_readout} and~\ref{eq:recurrence} gives the transformed update and readout:
\begin{equation}
\begin{aligned}
    \mT_t&=\mA'_t\mT_{t-1}+\beta_t\vk'_t{\vv'_t}^{\top},\\
    \vo'_t&=\mathbf D'_c\mT_t^{\top}\vq'_t=\mathbf P_V\vo_t.
\end{aligned}
\label{eq:transformed_recurrence}
\end{equation}
Multiplication by $\mathbf D'_c$ restores the numerical scale while keeping the value dimensions in the reordered coordinates. Normalization and output-gating parameters follow the same value order.
For this head, the output-projection weights become $\mathbf W'_O=\mathbf W_O\mathbf P_V^{\top}$, absorbing the inverse value permutation without an explicit reordering operation.
With the initial state transformed accordingly, the block is functionally equivalent to the original in the absence of quantization.

The fixed channel permutations are absorbed into the projection weights and associated channel-wise parameters before inference, so the recurrent state does not need to be reordered at each decoding step.
Value scaling and scale restoration are fused with the recurrent update and readout.

\subsection{Packed Layout and Fused Update}
\label{app:efficient_implementation}

\paragraph{State storage.}
The cache stores the transformed state $\mT_t$ from Equation~\ref{eq:permuted_state_coordinates} in two contiguous regions: FP16 values for the $\khi$ selected key channels and INT8 codes for the remaining channels.
Each group of 32 INT8 values along the value dimension has one FP32 quantization scale stored in the cache.
The contiguous regions support vectorized loads and stores.

\paragraph{Fused decoding.}
At each decoding step, a single CUDA kernel loads the packed state and dequantizes its INT8 codes.
It evaluates the transformed update and readout in Equation~\ref{eq:transformed_recurrence} using FP32 arithmetic, then
requantizes the low-precision channels with recomputed group scales.
The kernel writes back FP16 values, INT8 codes, and FP32 scales.
State tiles are reused in registers across these stages, reducing intermediate memory traffic.

\paragraph{Prefill path.}
Prefill uses FP32 arithmetic within each chunk and writes its final state to the same packed cache. 
Subsequent prefill chunks and decoding steps read this compressed state.

\subsection{Evaluation Settings}
\label{app:evaluation_settings}

\paragraph{Generation parameters.}
For downstream accuracy evaluation, Qwen3.6-35B and Kimi-Linear-48B run in thinking mode.
Qwen3.6 uses temperature $0.6$, top-$p=0.95$, top-$k=20$, and at most $65{,}536$ output tokens.  
Kimi-Linear uses temperature $1.0$, top-$p=1.0$, no top-$k$ truncation, and at most $262{,}144$ output tokens.
Kimi-K3 uses temperature $1.0$, top-$p=0.95$, zero presence and frequency penalties, and at most $262{,}144$ output tokens.

Table~\ref{tab:generation_counts} lists the number of responses generated for each problem on each benchmark.

\begin{table}[H]
\caption{Number of generations per problem for each model and benchmark.}
\label{tab:generation_counts}
\centering
\small
\setlength{\tabcolsep}{7pt}
\begin{tabular}{@{}lccc@{}}
\toprule
\textbf{Benchmark} & \textbf{Qwen3.6-35B} & \textbf{Kimi-Linear-48B} & \textbf{Kimi-K3} \\
\midrule
AIME 2026          & 128 & 128 & 16 \\
HMMT February 2026 &  64 &  64 &  8 \\
IMO-AnswerBench    &  16 &  16 & -- \\
GPQA-Diamond      &  32 &  32 & -- \\
MMLU-Pro          &  16 &  16 & -- \\
LiveCodeBench-v6   &   8 &   8 & -- \\
\bottomrule
\end{tabular}
\end{table}

\section{Long-Context Evaluation}
\label{app:long_context}

We evaluate long-context accuracy on RULER~\citep{hsieh2024ruler} at context lengths from 4K to 128K tokens, comparing FP32, FP16, uniform INT8+SR, and \method{} state storage (Table~\ref{tab:long_context}).
Each format is used throughout prefill and decoding. \begin{table}[H]
\caption{RULER macro accuracy (\%) across context lengths.}
\label{tab:long_context}
\centering
\small
\setlength{\tabcolsep}{9pt}
\begin{tabular}{lcccccc}
\toprule
\textbf{State format} & \textbf{4K} & \textbf{8K} & \textbf{16K} &
\textbf{32K} & \textbf{64K} & \textbf{128K} \\
\midrule
\rowcolor{GroupGray}\multicolumn{7}{l}{\emph{Qwen3.6-35B-A3B (GDN)}} \\
FP32 state            & 96.96 & 96.94 & 96.65 & 96.56 & 96.32 & 96.13 \\
FP16 state            & 96.98 & 96.94 & 96.65 & 96.56 & 96.32 & 96.11 \\
INT8+SR               & 96.93 & 96.89 & 96.60 & 96.43 & 96.28 & 96.07 \\
\rowcolor{DAMPBlueLight}
\textbf{\method{} INT8} & \textbf{97.00} & \textbf{96.94} & \textbf{96.65} &
\textbf{96.56} & \textbf{96.32} & \textbf{96.10} \\
\midrule
\rowcolor{GroupGray}\multicolumn{7}{l}{\emph{Kimi-Linear-48B-A3B-Instruct (KDA)}} \\
FP32 state            & 95.27 & 94.44 & 93.53 & 92.37 & 91.44 & 88.02 \\
FP16 state            & 95.27 & 94.44 & 93.53 & 92.37 & 91.44 & 88.03 \\
INT8+SR               & 95.27 & 94.35 & 93.41 & 92.24 & 91.36 & 87.87 \\
\rowcolor{DAMPBlueLight}
\textbf{\method{} INT8} & \textbf{95.27} & \textbf{94.42} & \textbf{93.53} &
\textbf{92.36} & \textbf{91.44} & \textbf{88.02} \\
\bottomrule
\end{tabular}
\end{table}

Both INT8+SR and \method{} remain close to FP32 across the evaluated context lengths on RULER.
The largest absolute accuracy differences between \method{} and FP32 are 0.04 percentage points on Qwen3.6 and 0.02 on Kimi-Linear, compared with 0.13 and 0.15 points, respectively, for uniform INT8+SR.

\section{Stability of Offline Signals}
\label{app:calibration_stability}

\subsection{Calibration-Split Agreement}

\noindent
\begin{minipage}[t]{0.56\linewidth}
\vspace{0pt}
We assess sensitivity to the calibration data by splitting the corpus into two disjoint, domain-balanced subsets and calibrating them separately using the score in Equation~\ref{eq:cumulative_decay_error} with $H=256$.
For each layer and head, we compare the selected top-16 key channels and the complete channel rankings.
The overlap between selected sets is $|\mathcal H_1\cap\mathcal H_2|/16$.
Averaged across layers and heads, it reaches 92.0\% for KDA and 93.2\% for GDN, compared with 12.5\% expected for random selection (Table~\ref{tab:calibration_split_stability}).
The complete rankings have mean Spearman correlations of 0.990 and 0.986, respectively.
\end{minipage}\hfill
\begin{minipage}[t]{0.40\linewidth}
\vspace{0pt}
\captionof{table}{Agreement between calibration splits, averaged across layers
and heads. Common is the number of shared top-16 channels.}
\label{tab:calibration_split_stability}
\centering
\scriptsize
\setlength{\tabcolsep}{2.2pt}
\renewcommand{\arraystretch}{1.05}
\begin{tabular}{@{}lccc@{}}
\toprule
\textbf{Arch.} & \textbf{Common} & \textbf{Overlap} & \textbf{Spearman} \\
\midrule
KDA    & 14.72 & 92.0\% & 0.990 \\
GDN    & 14.91 & 93.2\% & 0.986 \\
Random &  2.00 & 12.5\% & --    \\
\bottomrule
\end{tabular}
\end{minipage}

\subsection{GDN Decay Structure}

\begin{figure}[t!]
\centering
\includegraphics[width=\linewidth]{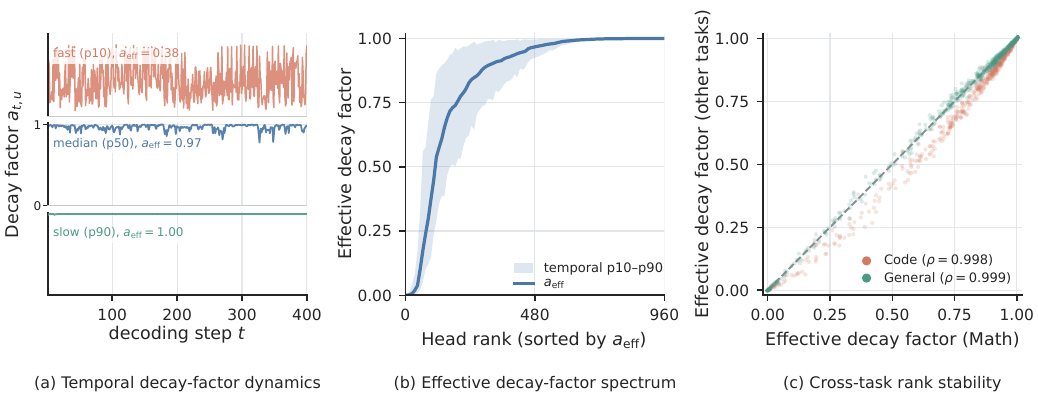}
\caption{Head-level decay structure in Qwen3.6-35B (GDN).
(a) Decay factors across decoding steps for heads at p10, p50, and p90 of
$a_{\mathrm{eff}}$.
(b) Effective decay factors across all 960 heads; shading shows the p10--p90
range across decoding steps.
(c) Effective decay factors on Code and General compared with Math, with
Spearman rank correlations.}
\label{fig:gdn_decay_structure}
\end{figure}

Figure~\ref{fig:gdn_decay_structure} summarizes decay statistics for all 960 heads in the 30 GDN layers of Qwen3.6-35B.
Since GDN shares a decay factor across all key channels within a head, we characterize each head by its effective decay factor, $a_{\mathrm{eff},\ell,h}=\exp(\mathbb E_t[\log a_{t,\ell,h}])$.
The effective decay factors at the 10th, 50th, and 90th percentiles are 0.376, 0.967, and 0.9996, respectively, indicating substantial differences in decay strength across heads.
Head rankings by effective decay factor are also highly consistent across tasks: the rankings on Code and General have Spearman rank correlations of 0.998 and 0.999, respectively, with those on Math.

\section{Additional GDN Ablations}
\label{app:gdn_ablations}

\paragraph{Key-channel selection.}
Table~\ref{tab:gdn_selector_appendix} compares selectors using the same INT8+SR quantizer and high-precision budget.
The Retention-only score is identical across key channels within a GDN head and cannot distinguish them.
In \method{}, however, the shared decay weight can vary across steps and weights each channel's local errors before averaging.
Thus, \method{} need not produce the same ranking as the error-only selector.

\begingroup
\setlength{\intextsep}{6pt}
\begin{table}[H]
\caption{Matched-budget GDN selector ablation at 9.9 bits per state value.}
\label{tab:gdn_selector_appendix}
\centering
\small
\setlength{\tabcolsep}{5pt}
\begin{tabular}{lccc}
\toprule
\textbf{Selector} & \textbf{AIME 2026} & \textbf{GPQA-D} & \textbf{LCB-v6} \\
\midrule
Random                    & 66.59         & 80.98         & 66.88         \\
State energy              & 82.19         & 81.37         & 85.17         \\
\rowcolor{DAMPBlueLight}
\textbf{\method{} ($C$)}
                          & \textbf{83.82} & \textbf{82.05} & \textbf{85.67} \\
\bottomrule
\end{tabular}
\end{table}
\endgroup

\section{Multi-Turn Serving on Kimi-K3}
\label{app:k3_multiturn_ttft}

\paragraph{Setting.}
Each configuration processes 256 four-turn WildChat-1M conversations, yielding 1,024 requests.
Output lengths match the corresponding dataset responses, capped at 1,024 tokens, and each generated response is included in the next turn's input.
Prefix caching is enabled in SGLang.
\method{} uses the main configuration with $\khi=16$ and FP32 group scales.
Table~\ref{tab:k3_multiturn_ttft} reports TTFT and the mean token-based cache hit rate across all four turns.

\section{Limitations}
\label{sec:limitations}

Our evaluation covers three released GDN- and KDA-based checkpoints in SGLang.  
Accuracy and efficiency gains may differ on other model architectures, hardware, and serving configurations.
Although \method{}-INT8 remains close to FP32, variants with INT4 or NVFP4 as the low-precision format do not yet recover FP32 accuracy.
Future work includes incorporating richer recurrent dynamics into channel selection, improving 4-bit state quantization, and evaluating these methods on additional model families and serving systems.

\end{document}